\documentclass{llncs}

\usepackage{amsmath,amssymb,amsfonts}
\usepackage{algorithmic}
\usepackage{graphicx}
\usepackage{textcomp}
\usepackage{color}
\usepackage{multirow}

\DeclareMathOperator*{\argmax}{arg\,max} 

\begin{document}

\title{SmartTennisTV: Automatic indexing of tennis videos}

\author{%
Anurag Ghosh \and 
C.V.Jawahar
}%
\institute{
CVIT, KCIS, IIIT Hyderabad\\
\email{anurag.ghosh@research.iiit.ac.in, jawahar@iiit.ac.in}
}

\maketitle

\begin{abstract}   
In this paper, we demonstrate a score based indexing approach for tennis videos. Given a broadcast tennis video (\textsc{btv}), we index all the video segments with their scores to create a navigable and searchable match. Our approach temporally segments the rallies in the video and then recognizes the scores from each of the segments, before refining the scores using the knowledge of the tennis scoring system. We finally build an interface to effortlessly retrieve and view the relevant video segments by also automatically tagging the segmented rallies with human accessible tags such as 'fault' and 'deuce'. The efficiency of our approach is demonstrated on \textsc{btv}'s from two major tennis tournaments. 
\end{abstract}

\section{Introduction}
\label{sec:intro}

Sports streaming websites are very popular with many services like TennisTV and WatchESPN offering full game replays on demand. Millions of users use these services for entertainment, education and other purposes. However, tennis matches are usually very long, often running into hours. It's very hard to infer playing styles and patterns of players without investing hundreds of hours of viewing time. Thus, it's cumbersome to find ``useful'' parts. Streaming websites provide the video as-is, i.e. it's only possible to access the video stream sequentially. However, in case of sports and other event-rich video streams, an useful extension is to provide random access (like accessing an array) grounded in events along with sequential access, so that extensions like skipping to next event, filtering events etc can be provided.

In this paper, we focus on constructing a point wise index of a tennis match and thus providing random access to the match. We propose a method to segment out the match into a set of rallies, then automatically extract the scorecard and the scores.Using tennis domain knowledge, we construct a novel algorithm to refine our extracted scores. We then demonstrate the utility of the automatically constructed index by building an interface to quickly and effortlessly retrieve and view the relevant point, game and set segments along with providing human accessible tags.  

There are multiple challenges in this scenario. The tennis match videos are recorded from multiple camera angles and edited to have different kind of shots, to capture various emotions and drama along with the game play. With respect to extracting scores, the score board is never at a fixed position or in a specific format and the score digits are not constrained by font, size, and color.

\begin{figure}[t]
\begin{center}
    \includegraphics[width=1\linewidth]{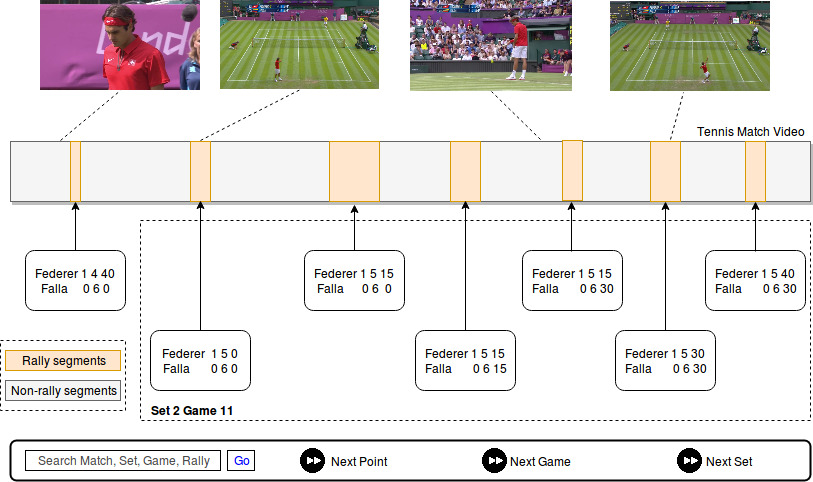}
\end{center}
   \caption{We aim to provide random access to tennis match videos and construct a point wise index of a tennis match so that a user can access, jump and skip ``points'', ``games'' and ``sets''.}
\label{fig:motivatingfigure}
\end{figure}

\begin{figure*}[t]
\begin{center}
\includegraphics[width=1\linewidth]{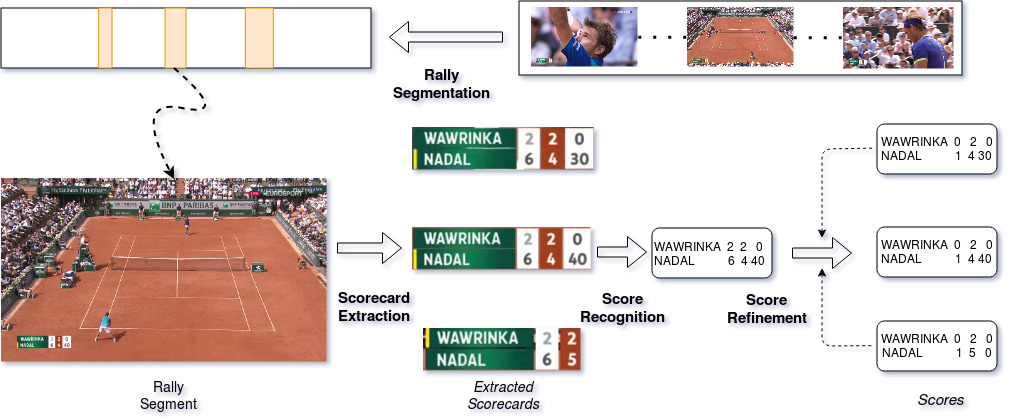}
\end{center}
   \caption{Our approach is illustrated in this figure. We start by temporally segmenting out the rallies, extracting the scoreboard and then recognizing the scores where we use contextual and domain knowledge to refine the recognized scores.}
\label{fig:pipeline}
\end{figure*}

The major contributions of this paper are, 
\begin{enumerate}
\item An effective score recognition algorithm using domain knowledge which can be adapted for different games. Here, we do our experiments on tennis videos by using the tennis domain knowledge.
\item We propose a score based indexing system, to navigate and retrieve segments from large volumes of video data with considerable ease.
\item Our method also enables many applications of indexing, we demonstrate one such application, human accessible event tagging.
\end{enumerate}

Section \ref{sec:relatedwork} discusses advances and related work in literature. Section \ref{sec:approach} forms the core of the paper, describing our core approach. Lastly, Section \ref{sec:results} provides a brief background of tennis and a high level description of our dataset(s), describes the implementation details and the experiments we performed along with obtained results.

\section{Related Work}
\label{sec:relatedwork}

\textbf{Sports Understanding:} Using domain specific cues, several researchers have previously worked on improving sports understanding (specially tennis), with strides made in video summarization and automatically generating highlights~\cite{hanjalic2003generic,  ghanem2012context, huang2009intelligent}, generating descriptions~\cite{sukhwani2015tennisvid2text} and automatically segmenting coarse temporal scenes~\cite{zhang2007personalized}, annotating players~\cite{yan2014automatic, mentzelopoulos2013active} and tracking the ball~\cite{yan2007all, zhou2015tennis}.

~\textbf{Sports Video Indexing and Applications:} Xu et al.~\cite{xu2008novel} and Miyamori et al.~\cite{miyamori2000video} focus on semantic annotations exploiting tennis domain knowledge to build retrieval systems based on positions and actions. Sukhwani et al.~\cite{sukhwani2016frame} proposed a dictionary learning method for frame level fine grained annotations for a given video clip, but their annotations are also computed at the level of actions, useful in the context of computing player statistics. Kolekar et al.~\cite{kolekar2015bayesian} use audio features to detect events in soccer scenes and generate highlights. Liu et al.~\cite{liu2009framework} perform mutlimodal analysis to generate tennis video highlights while Connaghan et al.~\cite{connaghan2011game} attempt to segment out game into point, game and set, however, perform no score keeping and use multiple cameras to perform the task. However, these methods do not attempt to robustly index point level information to enable retrieval from the point of view of a viewer. Our work differs from all of these as we attempt to annotate point level information for a match.

~\textbf{Scorecard and Score Extraction:} Liao et al.~\cite{liao2015research} focus only on detecting the scorecard while Miao et al.~\cite{miao2007real} focuses on both detection and extraction of scores, however the algorithm is specific for Basketball. Tesseract~\cite{smith2007overview} is the commonly used \textsc{ocr} pipeline to detect text from images and documents which have a plain background. Convolutional Recurrent Neural Network (\textsc{crnn})~\cite{shi2016end} is applicable for performing end-to-end scene text recognition while Textspot~\cite{gupta2016synthetic} introduces a Fully Convolutional Regression Network (\textsc{fcrn}) which performs end-to-end scene text detection and for recognition, uses the intermediary stage of the pipeline based on the lexicon-encoding CNN from Jaderberg et al.~\cite{jaderberg2014synthetic}.

\section{Approach}
\label{sec:approach}

Our goal is to automatically create an index for tennis videos. We begin by describing a method to automatically segment rallies. Then we detect and localize the scorecard in each of these rallies and recognize the text to abstract out the game score state to annotate the video with the accessible tags. An overview of our pipeline can be seen in Fig.~\ref{fig:pipeline}.

\subsection{Rally Segmentation}

Our method of segmenting out rallies stems from the observation that in \textsc{btv}'s, the camera is only overhead when the rally is in play and nowhere else. The background is mostly static after the serve begins, and remains the same till a player wins a point. \textsc{hog} features are appropriate in such a scenario, so we extract frames from the Segment Dataset, downscale them, and extract \textsc{hog} features. We then learn a $\chi$-squared kernel SVM to label a frame either as a rally frame or a non-rally frame. Then, we use this learned classifier to label each frame of the BTV as part of a rally or otherwise and smoothen this sequence using Kalman filter to remove any false positives/negatives to obtain the segmented rallies. 

\begin{figure*}
\begin{center}
   \includegraphics[width=1\linewidth]{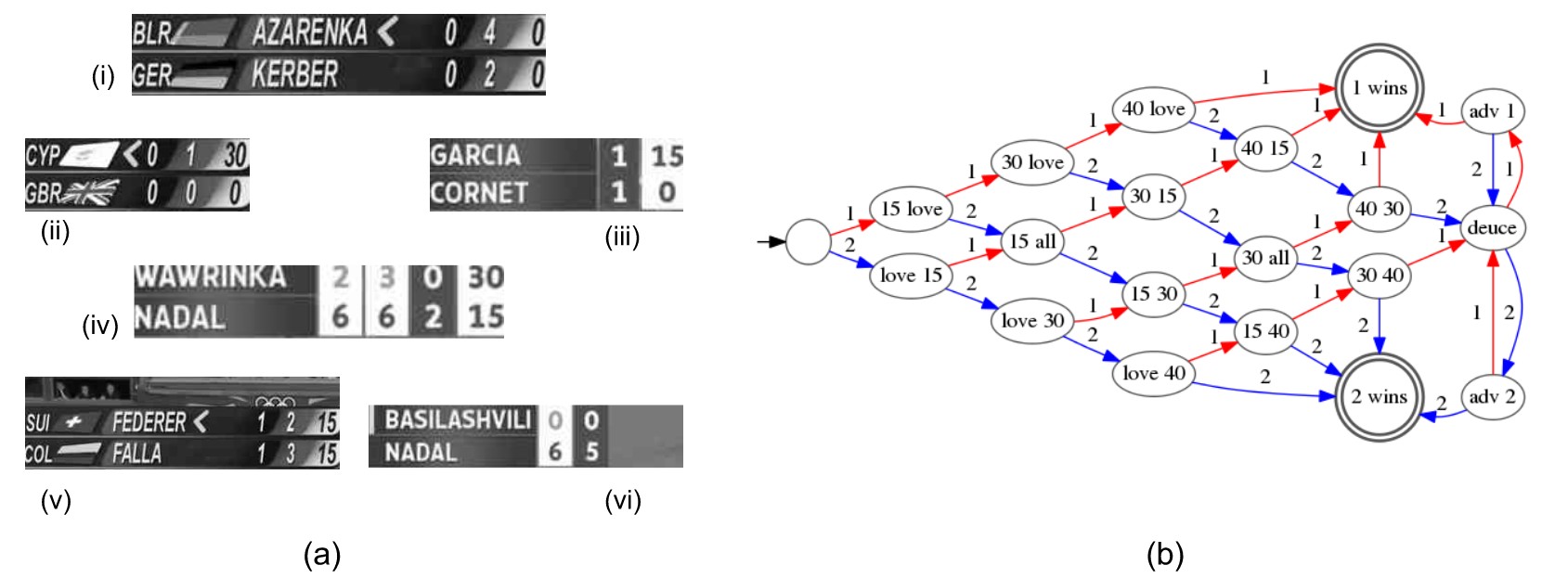}
\end{center}
   \caption{(a) depicts some of the extracted scorecards from different matches from our dataset. As one can see, the scorecards detected are of different sizes and formats, and differences across tournaments is noticeable. We have also included some of our failure cases, (v) and (vi) have extra regions that have been detected. (b) depicts the tennis point automaton that can be constructed from the tennis scoring system which is used to refine our extracted scores.}
\label{fig:prelimmontage}
\end{figure*}

\subsection{Scorecard Extraction}

We utilize the observation that the scorecard position is stationary in a rally, while the camera pans and moves around to cover the game. However, the scorecard may disappear and is not necessarily of the same size across the game as opposed to the assumptions in~\cite{miao2007real}. So, to overcome these issues, we extract the scorecard independently from each rally segment instead of assuming a single scorecard template.

We adapt the method described in~\cite{liao2015research}. We start by finding the gradient for each frame (say $I_{x}(i, j, t)$) using the sobel filter, and then calculate the normalized temporal sum for each frame using, $I_{norm}(i, j, n) =  \frac{1}{n} \sum_{t = 1}^{n} I_{x} (i, j, t)$. Then, we subtract $I_{x}$ and $I_{norm}$ to obtain the temporally correlated regions $I_{g}$. Further, we binarize the image using the following  equation,

\begin{equation}
I_{r}(i,j,t) = (1 - \frac{I_{x}(i,j,t)}{max_{t,i,j}(I_{x})})I_{norm}(i,j,t)
\end{equation}

Empirically, the scorecard is found in one of the corners of the frame, we identify the four regions of size $(h/5, w/2)$ in the corners as the regions to search for the scorecard. Note, $w$ and $h$ are the width and height of the frame respectively. We identify the coarse scorecard region by selecting the region with the maximum number of white pixels in the specified regions in $I_{r}(i,j,t)$ summed over time. Further, after we have identified the coarse region, we apply morphological operators to remove small aberrations present and fit a rectangle which encloses the scorecard area. Our qualitative results can be seen in  Fig.~\ref{fig:prelimmontage} {(a)}.

\subsection{Score Recognition}

Traditional \textsc{ocr} based methods like Tesseract~\cite{smith2007overview} can recognize text printed on a clear background however need the image to be preprocessed if the background is textured and shaded, and the contrast in the text fragments varies widely. However, with the advent of deep learning based \textsc{ocr} and scene text detection methods, a more general approach can be formulated. 

To recognize scores, we experiment with three different methods, Tesseract, \textsc{crnn} and Textspot. Textspot combines \textsc{fcrn}~\cite{gupta2016synthetic} which is an end to end text detection network, which constructs a field of predictors where each predictor is responsible for detecting a word if the word centre falls within the corresponding cell, akin to the \textsc{yolo} network architecture. The recognition is performed by the intermediary stage of the pipeline based on the lexicon-encoding \textsc{cnn} from Jaderberg et al~\cite{jaderberg2014synthetic}. \textsc{crnn}~\cite{shi2016end} is a scene text recognition network which treats the image as a sequence of strips. It proceeds by treating a \textsc{cnn} as a feature extractor to extract feature maps and construct a sequence of feature vectors. The sequence is fed into a bi-directional \textsc{lstm} to obtain label sequence probabilities and \textsc{ctc} loss is employed to obtain labels. We adapt and perform a comparison of the various score recognition baselines in Section~\ref{sec:results}. 

\subsection{Score Refinement}

To further refine our recognized scores, we use the knowledge of the tennis scoring system. As any structured game, score keeping in tennis is governed by a set of rules and thus, can be modeled as a finite automaton. Tennis in specific can be modeled as 3 automatons, one each for tracking the point, game and set score (See Fig.~\ref{fig:prelimmontage} (b)). Also, the vocabularies for point, game and set are restricted, so, we find errors by checking if the value belongs to the vocabulary or not. For instance, the the vocabulary for a point score is restricted to $\{ 0, 15, 30, 40, AD \}$.

Let $J = (game_{1}, set_{1}, point_{1}, game_{2}, set_{2}, point_{2})$ be the score state where game, set and point have the same meanings as in tennis. Firstly, we exploit the fact that the game and set scores are usually remain constant in a window, and thus replace errors with the mode of the value in the temporal window (with exceptions for score change within the window). 

Consider the tennis scoring automaton $T$ which is composed of score states and the transition function is constructed using the tennis scoring rules. Then we define a function $nextStates(s)$ which returns all possible states for the next game state. Likewise, $previousStates(s)$ provides the set of originating states for the current state $s$. For instance, from Fig.~\ref{fig:prelimmontage} (b), if we assume that we are at state $s = (0, 0, 30, 0, 0, 30)$ (referred to as 30 all in the figure), the function $previousStates(s)$ will return $\{ (0, 0, 30, 0, 0, 15), (0, 0, 15, 0, 0, 30) \}$ and $nextStates(s)$ would return $\{ (0, 0, 40, 0, 0, 30), (0, 0, 30, 0, 0, 40) \}$.

Assuming that the set of scores is $S = \{ s_{1}, s_{2} ... s_{n} \}$, and  that $s_{i}$ is erroneous (using vocabulary constraints), we compute the set $P = nextStates(s_{i-1}) \cap previousStates(s_{i+1})$, then we find the corrected score using,

\begin{equation}
s'_{i} = \argmax_{p \in P} \frac{1}{|J|} \sum_{j \in J} \delta(s_{i}(j), p_{i}(j))
\end{equation}

where $J$ is the set of game score states and $\delta$ is the Kronecker delta function. This equation is only needed if there are more than one possible score. It is to be noted that this method is extensible to any game which follows a structured scoring system like tennis.

\section{Experiments and Results}
\label{sec:results}

\begin{figure}[t]
\begin{center}
   \includegraphics[width=1\linewidth]{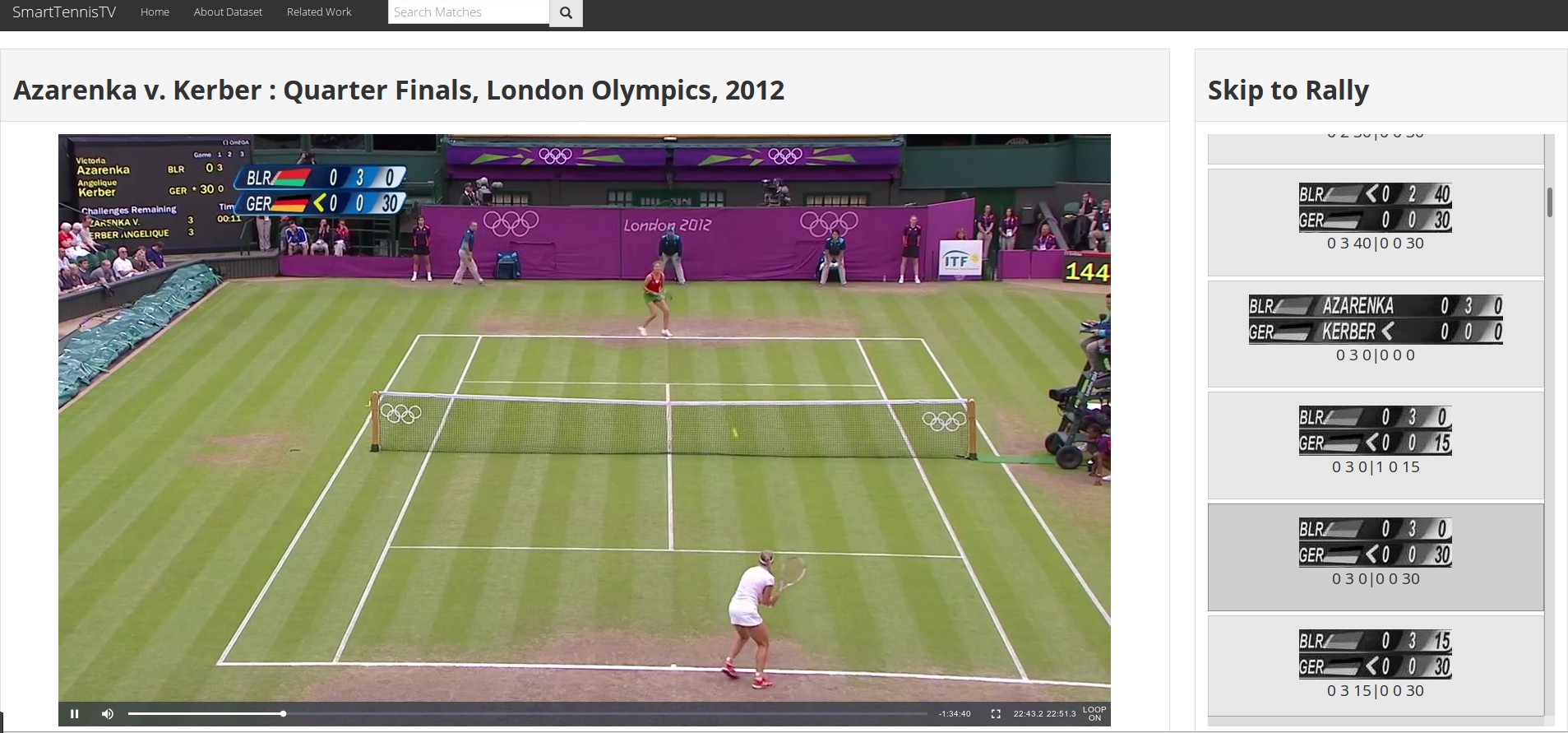}
\end{center}
   \caption{The developed interface supports the indexing and retrieval of a match as a point, game and set.}
\label{fig:coolimage}
\end{figure}

\subsection{Dataset}

A tennis match is divided into sets, each set is divided into games and each game has certain number of points or rallies. We restrict ourselves to ``singles'' matches and work with broadcast tennis video (\textsc{btv}) recordings at 720p for 10 matches. 5 matches are taken from the French Open 2017 and remaining matches are from London Olympics 2012 for all our experiments. For performing rally segmentation, we created a "Rally Dataset" by manually annotating 2 matches into rally and non rally segments. The training and test set images are derived by dividing all the images in a 50-50 split. For evaluating score extraction, we further annotated 4 matches with score of each segment using the automated segmented rallies from our described algorithm. All together, we have annotated 1011 rallies to create the "Match Scores Dataset".

\subsection{Rally Segmentation}

For learning the rally segmentation classifier, we extracted every 10th frame from Rally Dataset and cross validated using a 0.8 split to find the optimal values of the hyper-parameters $C$ and the period of the $\chi$-squared kernel. The optimal value of $C$ is $0.05$ and the period of the $\chi$-squared kernel \textsc{svm} is found to be $3$.

The mean $F1$ score on the test set for the task was found to be $97.46\%$, the precision for the non-rally segments was $98.94\%$ and the rally segments was $95.41\%$.

\subsection{Score Recognition}

\begin{table}
\begin{center}
\caption{Averaged Edit Distance for score recognition (Lower is better)}
\label{tab:algochooser}
\begin{tabular}{|l|l|l|l|l|}
\hline
Match & Textspot & \textsc{crnn} & Tesseract-P \\
\hline\hline
Match 1 (186 rallies) & 0.2070 & 0.4272 & 0.2612 \\
Match 2 (218 rallies) & 0.2178 & 0.4476 & 0.3780 \\
\hline
\end{tabular}
\end{center}
\end{table}

For employing Tesseract, we carefully preprocess the scorecard image and threshold the image manually. For each tournament such a preprocessing step needs to be manually defined. To train the \textsc{crnn}, which is constrained to recognize words as sequences, we divided the scorecard to two parts horizontally. For employing Textspot, we don't train the network and use the model trained on ``SynthText in the Wild'' dataset as~\cite{gupta2016synthetic} note state-of-the-art performance on standard benchmarks. However, we post-process the text detection boxes and sort them to extract the scores. We used edit distance instead of the usual text recognition metrics because the ``spaces'' between scores (in the recognized string) are relevant in our case. For instance, \textsc{crnn} removes repetitions of numbers, which causes the decrease in accuracy. Table~\ref{tab:algochooser} here presents our experimental results on a subset of the matches and as we can see, Textspot performed the best and thus, for the next set of experiments we use that as our baseline. 

\subsection{Score Refinement}

It is important to reiterate that our aim is not to recognize the text in the scorecard, but rather capture the game score state. To evaluate our results, we formulate a new metric, which inputs computed game state $C_{i}$ and the actual game state $G_{i}$, and computes the following (for a set of rallies say $R$),

\begin{equation}
AC(R) = \sum_{i \in R} \frac{1}{|J|} \sum_{j \in J} \delta(C_{i}(j), G_{i}(j))
\end{equation}

where J and $\delta$ as defined earlier.

\begin{table}
\caption{Averaged Score Accuracy AC(R) for our method and the defined baseline, \textsc{fcrn} (Higher is better)}
\label{tab:accresults}
\begin{center}
\begin{tabular}{|l|l|l|}
\hline
Match & Textspot & Ours \\
\hline\hline
Match 1 (186 rallies) & 79.30\% & 91.66\% \\
Match 2 (218 rallies) & 77.90\% & 80.58\% \\
Match 3 (201 rallies) & 92.45\% & 95.19\% \\
Match 4 (194 rallies) & 85.22\% & 92.18\% \\
\hline
\end{tabular}
\end{center}
\end{table}

As can be seen from Table~\ref{tab:accresults}, our refinement algorithm shows a consistent improvement in the averaged score accuracy across matches over the best performing baseline method, Textspot~\cite{gupta2016synthetic}. However, as it is apparent, the performance of our method is dependent on the performance of the baseline score recognition and that is possibly the reason in the relatively meager improvements in score accuracy in the second match.  

\subsection{Event Tagging}

\begin{table}
   \begin{center}
   \caption{Averaged Accuracy score of automatic event tagging (in percentage)}
   \label{tab:accesstags}
   \begin{tabular}{|l|l|l|l|l|l|l|l|l|}
    \hline
    \multirow{2}{*}{} &
      \multicolumn{2}{c}{Match 1} &
      \multicolumn{2}{c}{Match 2} &
      \multicolumn{2}{c}{Match 3} &
      \multicolumn{2}{c|}{Match 4} \\
    \hline
    & Textspot & Ours & Textspot & Ours & Textspot & Ours & Textspot & Ours \\
	\hline\hline
	Fault     & 66.66  & 70.83 & 52.24 & 56.71 & 87.87 & 90.90 & 84.44 & 84.44 \\
    Deuce     & 100.0 & 100.0 & 73.68 & 78.94 & 100.0 & 100.0 & 94.73 & 94.73 \\
    Advantage & 100.0 & 100.0 & 77.77 & 77.77 & 100.0 & 100.0 & 95.65 & 95.65 \\
	\hline
	Overall   & 75.00 & 79.41 & 60.58 & 64.43 & 92.45 & 94.33 & 89.65 & 89.65 \\
	\hline
  \end{tabular}
  \end{center}
\end{table}


Further, we automatically tagged common tennis events of importance to viewers such as ``fault'', ``deuce'' and ``advantage'' using simple rules which define these tennis terms and our extracted scores. We compare our accuracy with and without score refinement and can observe that there is an improvement corresponding to improvement in the score accuracy. Accuracy for each tag per match (the matches are same as Table~\ref{tab:accresults}) can be seen in Table~\ref{tab:accesstags}. 

\section{Conclusion}

We have presented an approach to create a tennis match index based on recognizing rallies and scores, supporting random access of ``points'' (Fig.~\ref{fig:coolimage}) tagged with common tennis events. Further extensions to this work are numerous, such as providing point based semantic search and performing tennis player analytics using videos instead of expensive sensor-based technologies.

\end{document}